\documentclass{article}

\usepackage[preprint]{neurips_2026}
\usepackage[utf8]{inputenc}
\usepackage[T1]{fontenc}
\usepackage{hyperref}
\usepackage{url}
\usepackage{booktabs}
\usepackage{amsfonts}
\usepackage{amsmath}
\usepackage{nicefrac}
\usepackage{microtype}
\usepackage{xcolor}
\usepackage[table]{xcolor}
\usepackage{array}
\usepackage{multirow}
\usepackage{arydshln}
\usepackage{graphicx}
\usepackage[table]{xcolor}
\usepackage{enumitem}
\usepackage{tabularx}
\usepackage{listings}
\usepackage[most]{tcolorbox}

\newtcblisting{promptbox}{
  colback=gray!4,
  colframe=gray!35,
  listing only,
  breakable,
  listing options={
    basicstyle=\ttfamily\scriptsize,
    breaklines=true,
    columns=fullflexible
  }
}

\newtcolorbox{mybox}[2][]
  {colback = black!5!white, colframe = black!75!black, fonttitle = \bfseries,
    colbacktitle = black!100!black, enhanced, before upper={\fontsize{10}{11}\obeyspaces\obeylines\selectfont}, fontupper=\selectfont,
    attach boxed title to top left={yshift=-2.2mm,xshift=2mm},
    title=#2,#1}

\title{What Makes Interaction Trajectories Effective for Training Terminal Agents?}

\author{%
\textbf{Sidi Yang}$^{1}$,
\textbf{Chaofan Tao}$^{2,\dagger}$,
\textbf{Jierun Chen}$^{2}$,
\textbf{Tiezheng Yu}$^{2}$,
\textbf{Ruoyu Wang}$^{3}$,
\textbf{Yuxin Jiang}$^{2}$,\\
\textbf{Yiming Du}$^{2}$,
\textbf{Wendong Xu}$^{1}$,
\textbf{Jing Xiong}$^{1}$,
\textbf{Taiqiang Wu}$^{1}$,
\textbf{Lifeng Shang}$^{2}$,
\textbf{Xiao-Hui Li}$^{2}$,\\
\textbf{Ngai Wong}$^{1}$,
\textbf{Haoli Bai}$^{2,\dagger}$\\[2mm]
$^{1}$The University of Hong Kong \quad
$^{2}$Huawei Technologies \quad
$^{3}$Nanyang Technological University\\
$^{\dagger}$Corresponding authors\\[1.5mm]
\small
\texttt{Project:} \href{https://stephen0808.github.io/terminal-lego.github.io/\#}
{https://stephen0808.github.io/terminal-lego.github.io/}
}

\begin{document}

\maketitle

\begin{abstract}
Stronger code agents are commonly assumed to be superior teachers for post-training, yet this assumption remains poorly disentangled from task difficulty, harness design, and student capacity. 
We investigate this pedagogical link using \textsc{Terminal-Lego}, a scalable pipeline that transforms multi-domain real-world issues into environment-verified agentic tasks.
Surprisingly, standalone performance does not dictate teaching efficacy: while Claude Opus 4.6 achieves higher scores on Terminal-Bench 2.0, students fine-tuned on trajectories from DeepSeek-V3.2, a lower-scoring agent, exhibit significantly stronger generalization.
We attribute this "pedagogical paradox" to \textit{Environment-Grounded Supervision} (EGS): trajectories that explicitly expose inspect-act-verify behaviors through harness-visible interactions allow students to internalize robust problem-solving routines rather than fragile action sequences.
Scaling analysis reveals exceptional data efficiency: with only 15.3k \textsc{Terminal-Lego} trajectories, for example, Qwen3-32B achieves a 24.3\% score on Terminal-Bench 2.0, rivaling previous SOTA performance established with over 30$\times$ the data volume.
Our results suggest that the frontier of agent post-training lies beyond mere outcome-matching, shifting the focus toward "Harness Engineering", where the systematic design of environment-grounded interaction structures serves as the primary catalyst for reproducible and generalizable agentic intelligence.
\end{abstract}

\begin{figure}[htp]
    \centering
    \includegraphics[width=0.95\linewidth]{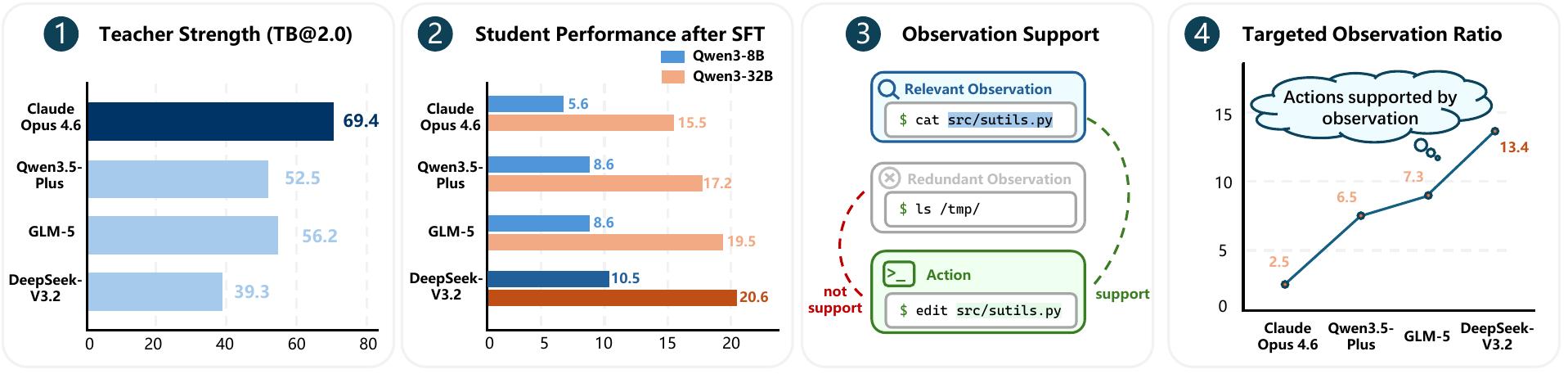}
    \caption{\textbf{The Pedagogical Paradox: Discrepancy between standalone performance and teaching efficacy}. While Claude Opus 4.6 achieves the highest standalone score on Terminal-Bench 2.0, its trajectories produce significantly weaker students compared to those from DeepSeek-V3.2. We attribute this gap to the alignment between actions and environmental feedback: teachers that prioritize actions rigorously supported by prior observations, a core property of \textit{Environment-Grounded Supervision}, provide robust, generalizable problem-solving routines that are more effective for student imitation learning.}
    \label{fig:placeholder}
\end{figure}

\section{Introduction}
Code agents are undergoing a fundamental paradigm shift from static code generation toward autonomous, closed-loop interaction with development environments \cite{merrill2026terminal, kwa2025measuring, xie2025swe, wang2025swe, jimenez2024swebench, badertdinov2025swerebench, Deng2025SWEBenchPC, yang2024swe}. In modern agentic workflows, exemplified by Cursor, Codex CLI \cite{openai2026codex}, and Claude Code \cite{anthropic2026claude}, a model is no longer judged solely by its final output, but by its ability to perceive complex environment states, execute interleaved actions, and iteratively verify outcomes. Terminal environments serve as a canonical testbed for this transition; they expose high-stakes skills such as dependency resolution, multi-file manipulation, and test-driven debugging through a unified, harness-mediated interface, offering a precise lens to study the mechanics of agentic reasoning.

This shift fundamentally redefines the objectives of post-training. An agentic problem-solving trajectory is no longer a monolithic response but a sequential trace of environmental grounding, capturing how an agent inspects, reflects, and adapts. Current distillation and fine-tuning practices typically operate on the "Stronger-is-Better" assumption: the stronger the teacher, the better the student. We challenge this notion by asking a critical, yet overlooked question: \emph{In the world of code agents, is a model's ability to solve a task truly the same as its ability to teach it?}

We study this question under controlled and realistic conditions. Existing terminal-agent data pipelines provide valuable but different substrates: TermiGen~\cite{zhu2026termigen} injects errors into generated tasks, TerminalTraj~\cite{wu2026large} mines executable repository trajectories, CLI-Gym~\cite{lin2026cli} constructs tasks through environment inversion, and Nemotron-Terminal~\cite{pi2026data} scales skill-based task synthesis. To isolate trajectory teachability across teachers, we construct \textsc{Terminal-Lego}: a scalable pipeline that extracts massive real StackOverflow issues and converts them into Docker-verified \emph{agentic terminal tasks}. Together with a fixed Terminus-2 harness, Terminal-Lego gives us a controlled substrate for comparing terminal-agent trajectories under the same task difficulty and interaction interface.

Our investigation uncovers a striking pedagogical paradox: standalone mastery does not guarantee teaching success. Under a matched-task setting, Claude Opus 4.6 \cite{anthropic2026int} achieves state-of-the-art (SOTA) performance as a standalone agent, yet its trajectories result in the least capable Qwen3 \cite{yang2025qwen3} students. Conversely, DeepSeek-V3.2 \cite{liu2025deepseek}, despite a lower standalone score, emerges as a superior teacher across both 8B and 32B student scales. This finding suggests that task-solving and knowledge-transfer are distinct, potentially orthogonal dimensions of agentic intelligence, where the "efficiency" of a teacher's solution may inversely correlate with its "teachability."

We trace this phenomenon to {Environment-Grounded Supervision} (EGS). We find that teachable trajectories are characterized by an explicit "inspect-act-verify" loop, making the internal reasoning process transparent through harness-visible interactions. While high-performing models often take "shortcuts" that minimize interaction, EGS-rich trajectories provide robust, generalizable problem-solving routines that allow students to internalize how to adapt rather than just what to output. To quantify this, we propose the Targeted Observation Ratio (TOR), a metric that measures the alignment between agent actions and environmental feedback, effectively predicting data utility prior to training.

The practical implications of our findings are substantial. By curating data based on interaction quality rather than sheer volume, we achieve exceptional data efficiency. Using only 15.3k \textsc{Terminal-Lego} trajectories, Qwen3-32B achieves a 24.3\% score on Terminal-Bench 2.0 (TB 2.0), a 7$\times$ improvement over its base performance, rivaling SOTA performance established with over 30$\times$ the data volume. Our results suggest that the frontier of agent post-training lies in "Harness Engineering", the systematic design of interaction structures as the primary catalyst for reproducible agentic intelligence.

Our contributions are threefold:
\begin{itemize}
\item \textbf{\textsc{Terminal-Lego} Agentic Data Pipeline}: We introduce a scalable pipeline that converts large-scale StackOverflow-grounded issues into Docker-verified tasks spanning 90+ domains, establishing a new standard for controlled, real-world agentic data synthesis.

\item \textbf{The Pedagogical Paradox \& EGS}: We identify a fundamental mismatch between agent performance and teachability, introducing Environment-Grounded Supervision (EGS) as a critical framework for curating effective post-training data.

\item \textbf{Targeted Observation Ratio (TOR)}: We propose and validate TOR as a predictive metric for trajectory quality, demonstrating that interaction-centric curation enables SOTA-level gains with unprecedented data efficiency (up to 30$\times$ less data than existing methods).

\end{itemize}

\section{Matched-Task Teacher Distillation}
\label{sec:study_setup}

This section defines the distillation setting used throughout the paper. Our goal is to compare the \emph{teachability} of trajectories rather than the raw problem-solving ability of teacher agents. To do so, we hold the task substrate, harness, student backbones, training recipe, and evaluation benchmark fixed whenever comparing teacher-generated trajectories.

We consider supervised fine-tuning (SFT) on multi-turn terminal-agent trajectories. Each training example records a teacher interacting with a Dockerized task environment through a fixed agent harness, Terminus-2~\cite{merrill2026terminal}. 
It operates through a single headless terminal inside a Docker container. At each turn, the model emits structured fields including \texttt{analysis}, \texttt{plan}, and shell \texttt{commands}; the harness executes the commands in a \texttt{tmux} session and returns captured terminal output. We use this fixed, model-agnostic harness for all teacher trajectory collection and student evaluation, so differences among trajectories primarily reflect teacher interaction behavior rather than scaffold-specific tools or model-specific agent engineering. Full Terminus-2 details are provided in Appendix~\ref{sec:terminus2}.

We use Terminal-Lego (Sec.~\ref{sec:terminal_lego}) to collect trajectories from four teacher models: DeepSeek-V3.2, Claude Opus 4.6, Qwen3.5-Plus \cite{qwen2026towards}, and GLM-5\cite{ZhipuAI2026GLM5}. To isolate trajectory quality from task difficulty, we focus on task-aligned subsets where all teacher models successfully solve the same instances, then train Qwen3-8B and Qwen3-32B students. We evaluate student performance on Terminal-Bench 2.0 and report average pass rate across three independent trials.


\section{Terminal-Lego: A Controlled Substrate from Real Terminal Issues}
\label{sec:terminal_lego}

A study of trajectory teachability requires tasks that are realistic enough to induce genuine terminal interaction, but controlled enough to support matched teacher comparisons. We therefore construct \textsc{Terminal-Lego}, a scalable pipeline that converts real user-facing technical issues into executable, Docker-verified agentic tasks.

\begin{figure}
    \centering
    \includegraphics[width=1.0\linewidth]{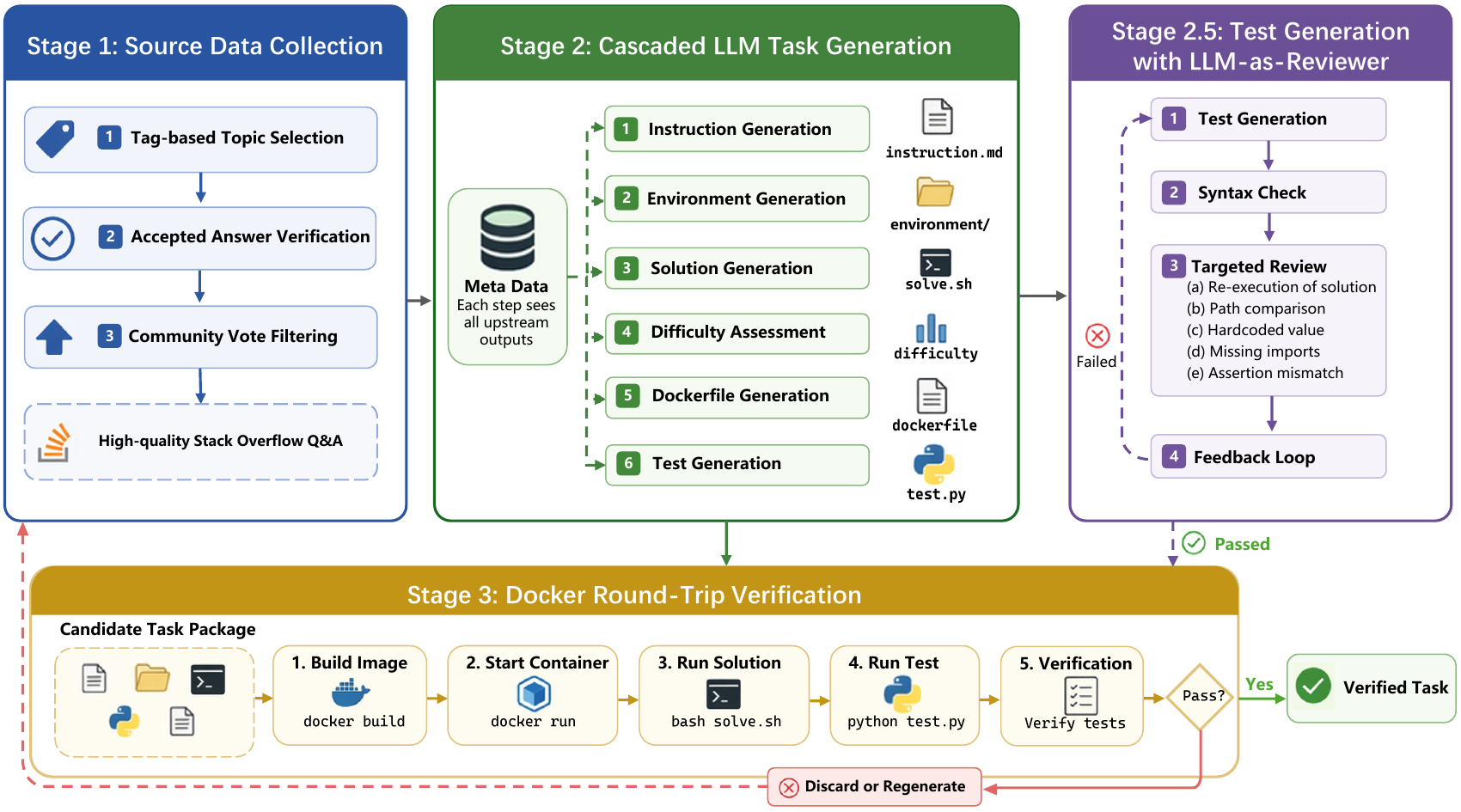}
    \caption{Terminal-Lego construction pipeline. StackOverflow issues are filtered into realistic sources, converted through cascaded task construction, and retained after Docker round-trip verification.}
    \label{fig:data-generation-pipeline}
\end{figure}

\subsection{Source Collection from StackOverflow}

We sample StackOverflow questions across 90+ technical domains. Each question is required to have an accepted answer, which provides a practical solution signal from the original asker. We further filter high-quality data by community vote thresholds.

This source distribution is useful for studying code agents because StackOverflow questions encode real failure modes: dependency conflicts, path mistakes, shell behavior, package installation problems, file-format conversions, networking configuration, and library-specific errors. These problems are broader than repository-centric software-engineering tasks and more grounded than purely synthetic skill templates. They also span diverse terminal-facing code-agent scenarios, making StackOverflow a scalable source for constructing agentic data that requires models to inspect, modify, execute, and verify real environment states.

\subsection{Cascaded Task Construction}

Each StackOverflow question is converted into a Terminal-Bench-style task through cascaded large language model (LLM) generation. The key design choice is that each stage conditions on upstream artifacts, making task construction a consistency problem rather than independent generation of unrelated files.


Because the instruction, environment, solution, Dockerfile, and tests are generated as a dependent chain, each retained task must describe one coherent executable terminal problem rather than a set of loosely related files.

\subsection{Test Review and Docker Round-Trip Verification}

LLM-generated tests can fail in systematic ways: they may rerun the solution, assume brittle paths, omit imports, hardcode inconsistent values, or assert properties that do not follow from the task. We therefore use a generate-then-review loop. Candidate tests are checked and reviewed by an independent LLM for common defect categories. Failed reviews are fed back into the next generation round.

Finally, every retained task must pass Docker round-trip verification. The validator builds the Docker image, runs the reference solution, executes the generated tests inside the container, and retains only tasks whose post-solution reward is positive. This full lifecycle is important for our trajectory study: teachers interact with tasks that are executable, automatically checkable, and comparable across models.
Additional details are provided in Appendix~\ref{app:pipeline-complete}.


\section{Why Stronger Teachers Can Teach Worse}

\subsection{A Pedagogical Paradox in Matched-Task Distillation}

We first test whether stronger benchmark models produce better SFT trajectories. A natural hypothesis is that stronger models should produce better training trajectories for student agents. Intuitively, models with higher task success rates are expected to generate more accurate and efficient interaction sequences, which should serve as higher-quality supervision signals during SFT. In our experiments, to eliminate potential biases arising from task variance, we curate 8.1k successfully passed trajectories from a common task set for each teacher and train the same student models.

\begin{table}[t]
  \caption{Performance of Qwen3-8B and Qwen3-32B across teacher-distilled trajectories. Although Claude Opus 4.6 has the highest standalone TB 2.0 score, DeepSeek-V3.2 produces the strongest student models.}
  \label{tab:teacher-selection}
  \centering
  \small
  \begin{tabular}{lcccccc}
    \toprule
    \multirow{2}{*}{Teacher Model} & \multirow{2}{*}{Trajectory Num.} & \multirow{2}{*}{TB 2.0} & \multirow{2}{*}{Avg. Turns} & \multicolumn{2}{c}{\textbf{SFT Avg. Pass@1}} \\
    \cmidrule(r){5-6}
    & & & & \textbf{Qwen3-8B} & \textbf{Qwen3-32B} \\
    \midrule
    Claude Opus 4.6 & 8.1K & 69.4 & 3.8 & 5.6\% & 15.5\% \\
    Qwen3.5-Plus & 8.1K & 52.5 & 6.7 & 8.6\% & 17.2\% \\
    GLM-5 & 8.1K & 56.2 & 5.4 & 8.6\% & 19.5\% \\
    DeepSeek-V3.2 & 8.1K & 39.3 & 7.3 & \textbf{10.5\%} & \textbf{20.6\%} \\
    \bottomrule
  \end{tabular}\vspace{-1em}
\end{table}

Surprisingly, results in table~\ref{tab:teacher-selection} contradict the intuition. Claude Opus 4.6 is the strongest standalone task solver in this group, yet its traces are the weakest imitation data. DeepSeek-V3.2 is the weakest standalone task solver, yet it produces the strongest students at both model scales.
This indicates that trajectory quality does not reflect the teacher benchmark score. We next rule out two simpler explanations--trajectory length and explicit error recovery--before isolating Environment-Grounded Supervision as the stronger mechanism.

\subsection{Are Trajectory Length and Error Recovery Truly Decisive?}
\label{Longest_shortest}

DeepSeek-V3.2 trajectories are longer on average than those of other teachers. One possible explanation is that longer traces contain more mistakes and recoveries, which may teach the student how to handle failures\cite{zhu2026termigen}. We test this explanation in two ways.

We identify 1.1k hard instances and generate five DeepSeek-V3.2 rollouts for each. Among successful attempts, we compare the shortest and longest successful trajectories for the same tasks. Longer trajectories contain more error turns and therefore serve as a proxy for higher recovery density.
Results are shown in Table~\ref{tab:ruling-out}: longest trajectories do not improve training and underperform the shortest successful trajectories. Based on the results, we find that simply extending trajectory length or introducing error recoveries may not effectively improve trajectory quality.

On the other hand, we further filter the 8.1k successfully passed trajectories by removing those whose terminal outputs contain error messages, yielding a 1.7k error‑free common set (the same tasks across all four teacher trajectory sets). However, even under this controlled setting, DeepSeek-V3.2 still produces the strongest student models. More importantly, compared to the full set in Table~\ref{tab:teacher-selection}, the student models trained from DeepSeek-V3.2 exhibit only a small performance degradation (1.5\% drop on Qwen3-32B), whereas all other teachers show a degradation of more than 5\%. This suggests that DeepSeek-V3.2, as a teacher, possesses an inherent consistency that insulates its teaching quality from task difficulty.

\begin{table*}[t]
  \caption{Ruling out trajectory length and explicit error recovery. Left: longer successful DeepSeek-V3.2 trajectories contain more error turns but produce weaker students than shortest successful trajectories. Right: DeepSeek-V3.2 remains the strongest teacher after filtering trajectories with explicit error messages.}
  \label{tab:ruling-out}
  \centering
  \small
  \setlength{\tabcolsep}{4pt}
  \renewcommand{\arraystretch}{1.08}
  \begin{minipage}[t]{0.47\textwidth}
    \centering
    \textbf{Longest vs. Shortest Successful Rollouts}\\[0.4em]
    \begin{tabular}{lcc}
      \toprule
      \textbf{Metric} & \textbf{Shortest} & \textbf{Longest} \\
      \midrule
      Avg. Turns & 8.9 & 12.2 \\
      Avg. Error Turns & 2.7 & 3.8 \\
      TOR & 15.5\% & 14.8\% \\
      \midrule
      Qwen3-8B & \textbf{6.0\%} & 5.2\% \\
      Qwen3-32B & \textbf{13.9\%} & 12.7\% \\
      \bottomrule
    \end{tabular}\vspace{-1em}
  \end{minipage}
  \hfill
  \begin{minipage}[t]{0.50\textwidth}
    \centering
    \textbf{Error-Free Teacher Comparison}\\[0.4em]
    \begin{tabular}{lcccc}
      \toprule
      \textbf{Metric} & \textbf{Claude} & \textbf{Qwen} & \textbf{GLM} & \textbf{DeepSeek} \\
      \midrule
      Traj. Num. & 1.7K & 1.7K & 1.7K & 1.7K \\
      Avg. Turns & 2.9 & 4.9 & 4.1 & 5.4 \\
      \midrule
      Qwen3-8B & 3.7\% & 5.6\% & 5.8\% & \textbf{7.9\%} \\
      Qwen3-32B & 10.5\% & 11.2\% & 15.4\% & \textbf{19.1\%} \\
      \bottomrule
    \end{tabular}\vspace{-1em}
  \end{minipage}
\end{table*}

These results suggest that the useful signal is not simply the presence of errors or recoveries, which is a reflection of their capability. The difference among teachers may instead arise from the inference behavior pattern.

\subsection{Environment-Grounded Supervision}

The standard view treats a trajectory as valuable when it solves the task. We propose a broader view: a terminal-agent trajectory is teachable when it exposes a reusable procedure for acting under environmental uncertainty. We call this property \textit{Environment-Grounded Supervision} (EGS): supervision in which the teacher makes observe-act behavior visible through harness-visible commands, terminal observations, and subsequent revisions. In terminal settings, EGS includes:
\begin{itemize}
    \item inspecting the initial filesystem, task constraints, dependencies, and runtime state;
    \item making state-changing actions such as editing files, installing packages, or running scripts;
    \item observing whether each action had the intended effect;
    \item adapting based on command output or environmental mismatch.
\end{itemize}

To quantify one aspect of this behavior, we define the \textit{Targeted Observation Ratio} (TOR) as a proxy for whether actions are supported by path-aligned prior observations. Let $\mathcal{A}$ denote the set of action commands in a trajectory, and let $\mathcal{O}$ denote the set of observation commands, including environment inspection and verification commands such as \texttt{cat}, \texttt{ls}, \texttt{find}, \texttt{grep}, \texttt{head}, \texttt{wc}, \texttt{diff}, and \texttt{stat}. For each action command $a \in \mathcal{A}$, we check whether there exists a prior observation command $o \in \mathcal{O}$ whose target path matches, contains, or is directly related to the target path of $a$. We define:
\begin{equation}
\mathrm{TOR} =
\frac{
|\{a \in \mathcal{A} : \exists o \in \mathcal{O},\ o \prec a \ \land\ \mathrm{align}(o, a)\}|
}{
|\mathcal{A}|
}.
\end{equation}

Here, $o \prec a$ indicates that observation $o$ occurs before action $a$, and $\mathrm{align}(o, a)$ indicates that the observation and action are path-aligned. For example, inspecting \texttt{src/utils.py} before editing \texttt{src/utils.py}, listing \texttt{src/} before creating a file inside it, or reading a script before executing it is counted as aligned support. In contrast, observing unrelated files or directories is not counted. Therefore, TOR measures the fraction of actions that are supported by relevant prior observations. 

We group observation commands by the kind of environment state they expose, as summarized in Table~\ref{tab:main-observation-taxonomy}. Appendix~\ref{sec:appendix_oa_patterns} provides the full command-level analysis.

\begin{table}[t]
\centering
\small
\caption{Observation command taxonomy used to interpret environment grounding in terminal-agent trajectories.}
\label{tab:main-observation-taxonomy}
\begin{tabular}{p{2.2cm}p{4.5cm}p{6.2cm}}
\toprule
\textbf{Category} & \textbf{Representative Commands} & \textbf{Environment State Exposed} \\
\midrule
Content & \texttt{cat}, \texttt{grep}, \texttt{head}, \texttt{wc}, \texttt{diff} & File contents, output text, checksums, and other evidence for whether produced data are correct. \\
Structure & \texttt{ls}, \texttt{find}, \texttt{stat}, \texttt{file}, \texttt{which} & Paths, directory layouts, file types, permissions, symlinks, and binary availability. \\
\bottomrule
\end{tabular}\vspace{-1em}
\end{table}

We find that DeepSeek-V3.2 differs from the other teachers in how frequently it observes intermediate state. Rather than issuing a compact sequence of direct edits, it often checks the filesystem, inspects generated files, runs partial commands, and confirms expected outcomes before proceeding. We next test whether this environment-grounded structure directly contributes to learning, then use failed trajectories and high/low-TOR subsets as supporting evidence that process quality is partly separable from final task success.

\begin{table*}[t]
\centering

\begin{minipage}[t]{0.45\textwidth}
  \centering
  \caption{Mechanism evidence for Environment-Grounded Supervision (EGS). Masking observation supervision in trajectories sharply reduces student performance.}
  \label{tab:mechanism-evidence}
  \small
  \begin{tabular}{lcc}
    \toprule
    \textbf{Metric} & \textbf{No Mask} & \textbf{Mask} \\
    \midrule
    Avg. Unmask Turns & 7.3 & 4.4 \\
    TOR & 13.4\% & 5.3\% \\
    \midrule
    Qwen3-8B & \textbf{10.5\%} & 3.8\% \\
    Qwen3-32B & \textbf{20.6\%} & 13.8\% \\
    \bottomrule
  \end{tabular}
  \vspace{-1em}
\end{minipage}
\hfill
\begin{minipage}[t]{0.53\textwidth}
  \centering
  \caption{Sensitivity to Targeted Observation Ratio (TOR) for Qwen3-32B. High-TOR trajectories outperform low-TOR trajectories and a random baseline at the same data scale.}
  \label{tab:vr-sensitivity}
  \small
  \vspace{0.8em}
  \begin{tabular}{lccc}
    \toprule
    \textbf{Metric}  & \textbf{High-TOR} & \textbf{Low-TOR} & \textbf{Random} \\
    \midrule
    Traj. Num. & 1.1K & 1.1K & 1.1K \\
    TOR & 14.3\% & 11.1\% & 13.5\% \\
    \midrule 
    Qwen3-8B & \textbf{7.5}\% & 5.6\% & 7.1\% \\
    Qwen3-32B  & \textbf{14.6}\% & 11.8\% & 13.8\% \\
    \bottomrule
  \end{tabular}\vspace{-1em}
\end{minipage}

\end{table*}

\subsubsection{Structured Failures Still Teach Students}

If EGS is a reusable interaction pattern rather than only a byproduct of final success, then even failed trajectories may contain positive pedagogical signal. We train students on 2.5k failed DeepSeek-V3.2 trajectories. These traces do not contain successful final solutions, so their main useful content is procedural: how the teacher observes, acts, and verifies.

\begin{table}[t]
  \caption{Passed and failed teacher trajectories comparison. Failed DeepSeek-V3.2 trajectories still train a surprisingly strong 32B student, suggesting that procedural EGS can transfer even without successful final states.}
  \label{tab:failed-trajectories}
  \centering
  \small
  \begin{tabular}{lccccc}
    \toprule
    \textbf{Teacher} & \textbf{Claude Opus 4.6} & \textbf{Qwen3.5-Plus} & \textbf{GLM-5} & \textbf{DeepSeek-V3.2} & \textbf{DeepSeek-V3.2} \\
    \midrule
    \rowcolor[gray]{0.95} \multicolumn{6}{l}{\textit{Data Complexity Metrics}} \\
    Correctness & Passed & Passed & Passed & Passed & Failed \\
    Traj. Num. & 8.1K & 8.1K & 8.1K & 8.1K & 2.5K \\
    TOR & 2.5\% & 6.5\% & 7.3\% & 13.4\% & 14.1\% \\
    \midrule
    \rowcolor[gray]{0.95} \multicolumn{6}{l}{\textit{SFT Performance (Avg Pass@1)}} \\
    Qwen3-8B   & 5.6\% & 8.6\% & 8.6\% & \textbf{10.5\%} & 3.5\% \\
    Qwen3-32B   & 15.4\% & 17.2\% & 19.5\% & \textbf{20.6\%} & 16.1\% \\
    \bottomrule
  \end{tabular}\vspace{-1em}
\end{table}

Table~\ref{tab:failed-trajectories} shows that the 8B student struggles with failed traces, as expected, but the 32B student remains competitive: with only 2.5k failed trajectories, it outperforms the model trained on 8.1k passed Claude Opus 4.6 trajectories. This suggests that sufficiently capable students can extract reusable interaction patterns from imperfect trajectories.

\subsubsection{High-Observation versus Low-Observation Rollouts}

We next compare high- and low-TOR trajectories from 5 rollouts by DeepSeek-V3.2 on the same hard-task pool mentioned in Sec.~\ref{Longest_shortest}. We train Qwen3-32B on 1.1k trajectories from each subset and compare against random selection. Note that the trajectories are sampled by observation ratio from the successfully passed trajectory pools and confirm all the tasks in three subsets are the same. Hence, this selection strategy isolates the task difficulty and diversity.

From Table~\ref{tab:vr-sensitivity}, the high-TOR subset produces a better student than the low-TOR counterpart. This supports the view that targeted observation ratio is associated with teachability, beyond teacher identity and task set.

\subsubsection{Observation Masking Supervision}

We next test whether observation commands are directly contribute to learning. We mask turns that only contain observe commands in 8.1k DeepSeek-V3.2 trajectories while preserving the remaining pure action or mix-command turns. Importantly, observation masking is implemented as a loss-level intervention rather than a trajectory-level deletion. The selected observation turns are still kept in the serialized trajectory and remain visible in the context for subsequent turns, together with their corresponding terminal observations.

Masking observation commands causes large drops for both student scales, as shown in Table~\ref{tab:mechanism-evidence}. This provides direct evidence that explicit observation supervision is not merely redundant checking. Students benefit not only from seeing observation results in context, but from being directly supervised to actively generate observation actions themselves.

\subsubsection{Targeted Observation Masking}
Observations consist of two components: targeted and untargeted. To test whether each component carries a learnable signal, we mask 50\% of targeted observations and the same number of turns of untargeted observations in DeepSeek-V3.2 trajectories. We then fine‑tune Qwen3‑32B against a random observation masking baseline, controlling for the total number of masked turns. Note that a turn may contain both action and observation commands; we only mask turns that contain observation commands, since decoupling actions from turns with mixed commands is difficult.

\begin{table*}[t]
  \caption{ Targeted and untargeted observation masking on Qwen3-32B. Compared with random masking, masking targeted observation turns leads to larger performance degradation under a matched masking budget.}
  \label{tab:mask-sensitivity}
  \centering
  \small
  \setlength{\tabcolsep}{4pt}
  \renewcommand{\arraystretch}{1.08}
  \begin{tabular}{lcccc}
    \toprule

    \rowcolor[gray]{0.95} \multicolumn{5}{l}{\textit{Data Complexity Metrics}} \\
    Masking Strategy & None & Random Mask &  Mask Targeted &  Mask Untargeted \\
    Traj. Num. & 8.1K & 8.1K & 8.1K&  8.1K \\
    Avg. Unmask Turns & 7.3 & 6.5 & 6.5 & 6.5  \\
    TOR & 13.4\%  & 12.8\%  & 10.8\%  &  13.4\% \\
    \midrule
    \rowcolor[gray]{0.95} \multicolumn{5}{l}{\textit{SFT Performance (Avg Pass@1)}} \\
    Qwen3-32B   & \textbf{20.6\%} & 19.1\% & 17.2\%&  19.7\%\\
    \bottomrule
  \end{tabular}\vspace{-1.5em}
\end{table*}

Table~\ref{tab:mask-sensitivity} shows that both observation turns contribute to learning. When the number of unmasked turns is controlled, masking untargeted observations only mildly reduces performance from 20.6\% to 19.7\%. In contrast, random masking under the same turn budget reduces performance to 19.1\%, and masking targeted observations reduces performance further to 17.6\%. This indicates that two categories of observation contribute to the quality of trajectories, but the targeted observation turns contain task-relevant evidence that helps the student learn how actions should be grounded in the environment.

\subsubsection{Activating Observation Behavior}

To further examine the role of observation in interaction trajectories, we inject the following instructions into the system prompt to encourage more environment-grounded behavior:

\begin{small}
\begin{promptbox}
- Inspect the environment and relevant files first, as thoroughly as possible.
- Only begin write/create/modify actions after you have fully verified the current file contents and directory structure.
\end{promptbox}
\end{small}

Using the same teacher model, Claude Opus 4.6, and the common 8.1K instance, we generate another trajectories set. We find that TOR increases from 2.5\% to 6.6\%. This increase is accompanied by a substantial improvement in 32B student performance, from 15.4\% to 19.5\%. These results suggest that observation behavior remains an important factor in training effective terminal agents.

We also observe an additional effect at inference time. With the modified prompt, the pass rate of Claude Opus 4.6 on 15k Terminal-Lego instances increases from 88.7\% to 95.4\%. This indicates that encouraging  environment-grounded behavior can improve not only the quality of training trajectories, but also the teacher model's own task-solving performance during inference.

\begin{mybox}[colback=gray!10]{Takeaways}
    \begin{itemize}[leftmargin=2ex]
    \item  A teacher's raw capability does not directly determine teaching quality. Trajectory length and error-recovery loops are not decisive factors either.
    \item EGS captures a fundamental reasoning behavior that students must acquire to act reliably. It benefits both inference and training.
    \item Students can learn useful behavioral patterns from high targeted observation ratio trajectories even when those trajectories fail the task, decoupling process quality from outcome.
    \end{itemize}
\end{mybox}

\section{Main Result}\label{sec:main_exp}

\subsection{Post-Training Setup and Hyperparameters}

We train Qwen3-8B and Qwen3-32B for the final results. We use a learning rate of $1e^{-5}$, maximum sequence length of 40,960 tokens, global batch size 8, micro-batch size 1 per GPU, AdamW with $\beta=(0.9,0.95)$, cosine learning-rate scheduling with 5\% warmup, and gradient clipping at 1.0. Qwen3-8B is trained for 7 epochs, while Qwen3-32B is trained for 3 epochs. The final training set contains 15k DeepSeek-V3.2 trajectories on Terminal-Lego tasks, including 12.8k successful trajectories and 2.5k failed trajectories.

\subsection{Terminal-Bench 2.0 Performance}

\begin{table*}[h]
\centering
\caption{Compact Terminal-Bench 2.0 comparison. Official references use their submitted scaffolds; Terminal-Lego student rows use Terminus-2. Terminal-Lego uses 15k trajectories, substantially improves the Qwen3 backbones, and is competitive with several official leaderboard references despite using much smaller student models.}
\label{tab:terminal-bench-results}
\small
\setlength{\tabcolsep}{4pt}
\renewcommand{\arraystretch}{1.08}
\begin{tabularx}{\textwidth}{>{\raggedright\arraybackslash}X c >{\centering\arraybackslash}p{2.6cm} >{\centering\arraybackslash}p{2.0cm} >{\centering\arraybackslash}p{1.9cm}}
\toprule
\textbf{Model} & \textbf{Size} & \textbf{Post-Training Data} & \textbf{Scaffold} & \textbf{TB 2.0} \\
\midrule
\multicolumn{5}{l}{\textit{Official leaderboard references}} \\
GPT-5.5\cite{openai2026int55} & -- & -- & Codex CLI & 82.0 \\
Claude Opus 4.6\cite{anthropic2026int} & -- & -- & Terminus-2 & 62.9 \\
GLM 5\cite{ZhipuAI2026GLM5} & 744B & -- & Terminus-2 & 52.4 \\
DeepSeek-V3.2\cite{liu2025deepseek} & 685B & -- & Terminus-2 & 39.6 \\
GPT-5-Mini\cite{openai2025int5} & -- & -- & Terminus-2 & 24.0 \\
Qwen 3 Coder 480B\cite{qwen32025coder} & 480B & -- & Terminus-2 & 23.9 \\
Grok 4\cite{xai2025grok4} & -- & -- & Terminus-2 & 23.1 \\
GPT-OSS-120B\cite{openai2025intoss} & 120B & -- & Terminus-2 & 18.7 \\
GPT-OSS-20B\cite{openai2025intoss} & 20B & -- & Terminus-2 & 3.1 \\
\midrule
\multicolumn{5}{l}{\textit{Backbones and post-trained student models}} \\
Qwen3-8B\cite{yang2025qwen3} & 8B & -- & Terminus-2 & 2.5 \\
TerminalTraj-Qwen2.5-Coder-7B\cite{wu2026large} & 7B & 50.7K & Terminus-2 & 10.1 \\
Nemotron-Terminal-Qwen3-8B\cite{pi2026data} & 8B & 490.5K & Terminus-2 & \textbf{13.0} \\
\rowcolor{blue!6} Terminal-Lego-Qwen3-8B & 8B & 15.3K & Terminus-2 & 11.8 \textcolor{blue}{\textbf{(+9.3)}} \\
\hdashline
Qwen3-32B\cite{yang2025qwen3} & 32B & -- & Terminus-2 & 3.4 \\
TermiGen-Qwen3-32B\cite{zhu2026termigen} & 32B & 3.5K & Bashagent & 18.0 \\
LiberCoder-Qwen3-32B\cite{lin2026cli} & 32B & 49.5K & OpenHands & 19.5 \\
TerminalTraj-Qwen2.5-Coder-32B\cite{wu2026large} & 32B & 50.7K & Terminus-2 & 22.0 \\
Nemotron-Terminal-Qwen3-32B\cite{pi2026data} & 32B & 490.5K & Terminus-2 & \textbf{27.4} \\
\rowcolor{blue!6} Terminal-Lego-Qwen3-32B & 32B & 15.3K & Terminus-2 & 24.3 \textcolor{blue}{\textbf{(+20.9)}} \\
\bottomrule
\end{tabularx}
\end{table*}

Table~\ref{tab:terminal-bench-results} compares Terminal-Lego with representative official Terminal-Bench 2.0 leaderboard entries, Qwen3 backbones, and prior post-training datasets. We keep a small number of recent high-scoring systems as upper-bound references and include several official calibration references near or below Terminal-Lego-Qwen3-32B to make the scale of the gains interpretable. Rows use different scaffolds in some prior systems, so cross-row comparisons should be read as broad references rather than a perfectly harness-matched leaderboard.

Using only 15k trajectories, Terminal-Lego-Qwen3-8B improves from 2.5\% to 11.8\%, clearly above the official GPT-OSS-20B reference. Terminal-Lego-Qwen3-32B improves from 3.4\% to 24.3\%, slightly exceeding official Terminus-2 results for GPT-5-Mini, Qwen 3 Coder 480B, and Grok 4. Our models do not surpass the models trained with the largest Nemotron-Terminal corpus, but we only use roughly 3\% of the reported data volume and approach its performance.

\subsection{Same-Size Comparison with Nemotron-Terminal}
For a fair comparison against Nemotron-Terminal, we sampled three subsets from skill-based data of its Terminal-Corpus with the same size as ours. By keeping the teacher model (DeepSeek-V3.2) fixed across both corpora, table~\ref{tab:nemotron-samples} shows that Terminal-Lego outperforms all three same-size Nemotron-Terminal subsets.

\begin{table}[t]
  \caption{Performance of Qwen3-32B across Terminal-Lego and three same-size subsets from Nemotron-Terminal. }
  \label{tab:nemotron-samples}
  \centering
  \small
  \begin{tabular}{lccc}
    \toprule
    Dataset & Trajectory Num.  & Avg. Turns & \textbf{SFT Avg. Pass@1} \\
    \midrule
    \rowcolor{blue!6} Terminal-Lego & 15K  & 8.2 & \textbf{24.3\%} \\
    Nemotron-Terminal-sample-1 & 15K & 8.0 & 19.9\% \\
    Nemotron-Terminal-sample-2 & 15K & 7.9 & 17.2\% \\
    Nemotron-Terminal-sample-3 & 15K & 8.0 & 17.9\% \\
    \bottomrule
  \end{tabular}\vspace{-2em}
\end{table}






\section{Related Work}

\subsection{Code Agents}
Recent progress has produced agentic systems that interact directly with software environments. Closed-source agents such as Codex\cite{openai2026codex}, Claude Code\cite{anthropic2026claude}, and Gemini CLI\cite{deepmind2025geminicli} combine long-context reasoning with tool and environment access. Open frameworks such as SWE-agent\cite{yang2024swe}, OpenHands\cite{wang2024openhands}, Mini-SWE-Agent\cite{miniswe2025miniswe}, and MiniSWE-Agent-Plus\cite{miniswe2025minisweplus} expose repo-level editing and execution through controlled action interfaces. Terminus and Terminus-2\cite{merrill2026terminal}, provide a particularly useful research substrate because every command and observation is visible through a compact harness. This visibility lets us study not only whether an agent solves a task, but how its interaction process can be learned by another model.

\subsection{Agentic  Coding Data Pipeline}
A growing line of work constructs data pipelines to train and evaluate interactive agents. In the realm of Software Engineering (SWE), frameworks such as SWE-rebench \cite{badertdinov2025swerebench}, SWE-smith \cite{yang2025swe}, R2E-Gym \cite{jain2025r2e}, SWE-Mirror \cite{wang2025swe} and SWE-Lego \cite{tao2026swe} have pioneered the conversion of repository-level issues into reproducible execution environments. Beyond pure SWE, the scope of code agents is rapidly expanding toward general computer-use. Terminal agents serve as the canonical interface for this broader paradigm; they encapsulate not only repository-level coding but also system administration, dependency resolution, execution debugging, and data processing workflows.
To support these diverse terminal-centric skills, recent pipelines such as TermiGen \cite{zhu2026termigen}, TerminalTraj \cite{wu2026large}, CLI-Gym \cite{lin2026cli}, and Nemotron-Terminal \cite{pi2026data} emphasize error-injected generation, repository mining, and skill-based synthetic scaling. Nevertheless, these prior works primarily optimize for task realism, corpus scale, or standalone agent robustness.

\section{Conclusions}
We investigated whether an agent’s standalone prowess dictates its pedagogical efficacy. Through the \textsc{Terminal-Lego} framework, we identified a "pedagogical paradox": a model’s task-solving capacity is not a reliable proxy for its teachability. We attribute this discrepancy to \textit{Environment-Grounded Supervision} (EGS), demonstrating that student learning is driven by the visibility of reasoning processes operationalized via our \textit{Targeted Observation Ratio} (TOR), rather than simple outcome-matching. Our results advocate for shifting the post-training focus from success rates to "interactive structures", the systematic loop of inspection, verification, and adaptation. By prioritizing these environment-grounded signals, we achieve state-of-the-art gains with 30$\times$ greater data efficiency than existing methods. We propose treating environment interaction as first-class supervision, positioning "Harness Engineering" as a pivotal frontier for developing robust, autonomous agents.

\subsection*{Limitations and Future Work}
\label{app:limit}
While our work advances the understanding of trajectory teachability for code agents and provides a high-quality executable dataset, we acknowledge several limitations that offer promising directions for future research, without undermining the core contributions of this study.

First, our experimental evaluations focus on 8B and 32B student models trained on trajectories generated by four representative teacher agents, covering a range of model scales and capabilities commonly used in current terminal-agent research. However, the generalizability of our key findings remains to be explored across a broader spectrum of model sizes. We anticipate that scaling effects or architectural differences may introduce nuanced variations in trajectory teachability. 

Furthermore, our TOR metric and masking analyses effectively capture critical aspects of observation usefulness, providing practical and interpretable evidence for the role of observational information in effective guidance signals (EGS). That said, these analyses do not fully address three finer-grained questions: whether a specific observation is strictly necessary for justifying the corresponding action, whether it is sufficient to support the action independently, and whether the teacher’s reasoning chain from observation to action is logically rigorous. As such, our measurements should be viewed as a practical characterization of EGS, rather than a comprehensive assessment of trajectory teachability, and future work could incorporate more formal reasoning validation to address these gaps. 

\noindent \textbf{Acknowledgements} This work was supported in part by the Theme-based Research Scheme (TRS) project T45-701/22-R of the Research Grants Council of Hong Kong, and in part by the AVNET-HKU Emerging Microelectronics and Ubiquitous Systems (EMUS) Lab.

\bibliography{nips26}
\bibliographystyle{nips26}

\appendix
\clearpage

\section{Terminal-Lego Pipeline: Complete Technical Details}
\label{app:pipeline-complete}

This appendix provides complete, reproducible details of the Terminal-Lego construction pipeline, including verbatim prompt templates, the Docker round-trip validation protocol. The complete source code will be available upon acceptance of this paper.

\subsection{Pipeline Overview}

The pipeline consists of six phases executed sequentially:

\begin{enumerate}
    \item \textbf{StackOverflow Crawling} -- Collect $\sim$36k questions with accepted answers across 98 weighted tags.
    \item \textbf{Cascaded Task Generation}  -- LLM generates 7 files per task in dependency order.
    \item \textbf{Docker Round-Trip Validation}  -- Build $\rightarrow$ solve $\rightarrow$ test $\rightarrow$ check reward.
    \item \textbf{Dataset Packaging} -- Renumber passed tasks into final corpus.
    \item \textbf{Oracle Evaluation} -- Sanity-check task solvability with oracle agent.
    \item \textbf{Teacher Evaluation} -- Evaluate each teacher model on the final task set.
\end{enumerate}

\subsection{Phase 1: StackOverflow Crawling}

\begin{figure}
    \centering
    \includegraphics[width=0.8\linewidth]{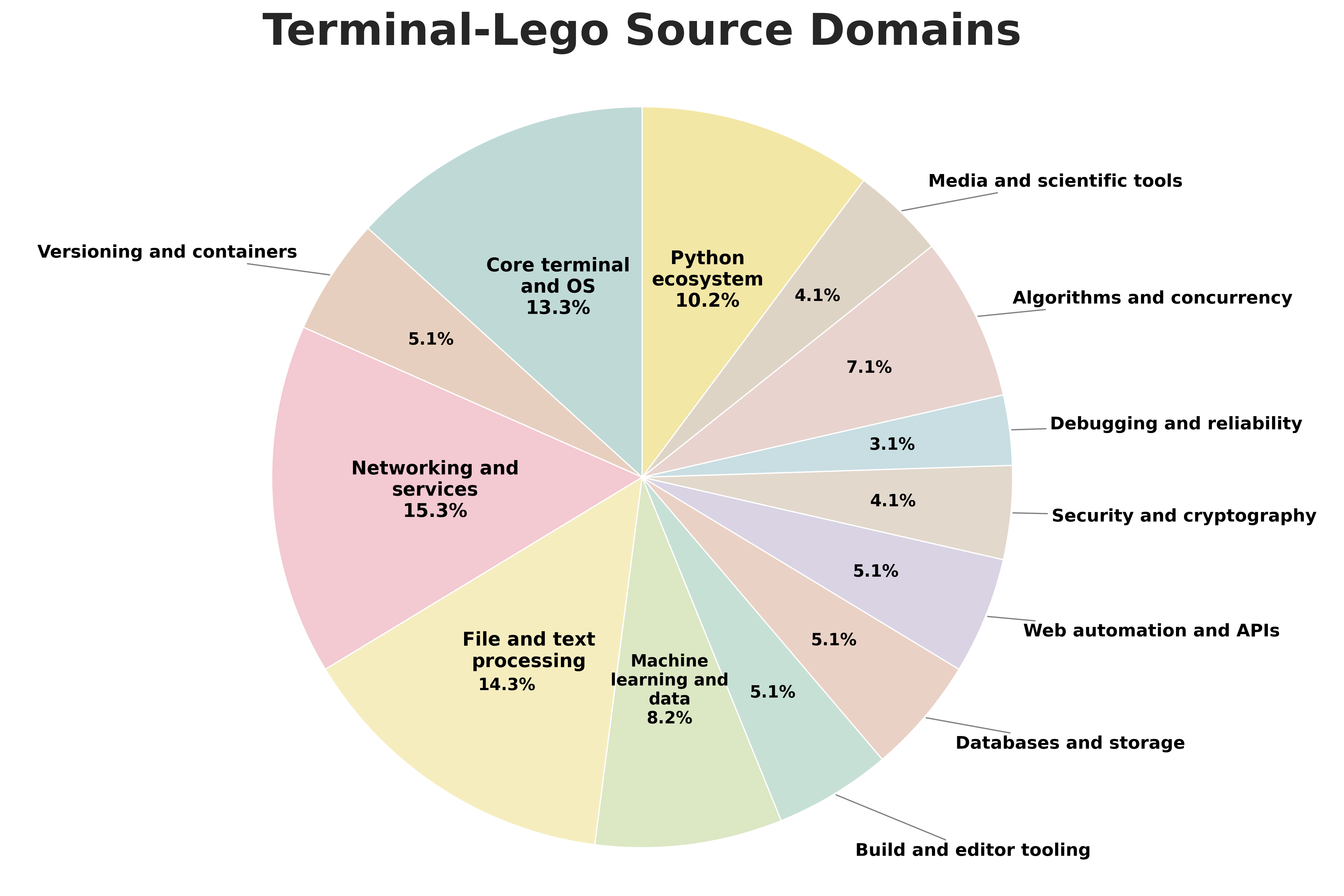}
    \caption{Distribution of Stack Overflow Source Questions Across Domains}
    \label{fig:appendix_pie_chart}
\end{figure}

The StackOverflow crawler uses 98 manually weighted tags chosen to approximate the skill distribution required by terminal-based tasks (Table~\ref{tab:appendix-domain-tags} and figure~\ref{fig:appendix_pie_chart}). Each tag is assigned a sampling weight, and the crawler allocates a target number of questions according to
\[
\texttt{tag\_target} = \min\left(\left\lfloor \frac{\texttt{weight}}{\sum_j \texttt{weight}_j} \cdot \texttt{TARGET\_COUNT} \cdot 1.5 \right\rfloor, \texttt{remaining}\right),
\]
where the 1.5 multiplier compensates for duplicate questions across tags. We require each question to contain an accepted answer, and we deduplicate by StackOverflow question ID against both historical crawls and the current crawl.

\begin{table}[h]
\centering
\small
\caption{Cascaded task construction in Terminal-Lego. Each stage conditions on upstream artifacts to keep the instruction, environment, solution, Dockerfile, and tests mutually consistent.}
\label{tab:cascaded-construction}
\begin{tabular}{p{2.35cm}p{5.0cm}}
\toprule
\textbf{Stage} & \textbf{Output and Role} \\
\midrule
Instruction & Executable task with paths and success criteria \\
Environment & Required input files and directory structure \\
Solution & Reference \texttt{solve.sh} conditioned on the generated environment \\
Difficulty & Easy/medium/hard label for analysis and sampling \\
Dockerfile & Runtime image and dependency setup \\
Tests & Post-solution checks over final environment state \\
\bottomrule
\end{tabular}
\end{table}

\textbf{Output:} A JSON file contains a metadata wrapper and a list of question objects. Each object contains: \texttt{question\_id}, \texttt{title}, \texttt{body}, \texttt{tags}, \texttt{score}, \texttt{accepted\_answer\_id}, \texttt{accepted\_answer} (with body and score), \texttt{link}, \texttt{search\_tag}, \texttt{creation\_date}.

\begin{table}[h]
  \caption{Domain grouping of StackOverflow tags used for Terminal-Lego source collection. Tags are manually grouped for readability; the crawler samples from the full tag list with per-tag weights.}
  \label{tab:appendix-domain-tags}
  \centering
  \small
  \begin{tabularx}{\textwidth}{p{3.2cm}X}
    \toprule
    \textbf{Domain} & \textbf{Tags} \\
    \midrule
    Core terminal and OS & \texttt{linux}, \texttt{bash}, \texttt{shell}, \texttt{command-line}, \texttt{terminal}, \texttt{unix}, \texttt{ubuntu}, \texttt{debian}, \texttt{centos}, \texttt{systemd}, \texttt{cron}, \texttt{sudo}, \texttt{chmod} \\
    Versioning and containers & \texttt{git}, \texttt{docker}, \texttt{kubernetes}, \texttt{virtualization}, \texttt{qemu} \\
    Networking and services & \texttt{networking}, \texttt{tcp}, \texttt{http}, \texttt{https}, \texttt{dns}, \texttt{ssl}, \texttt{ssl-certificate}, \texttt{openssl}, \texttt{curl}, \texttt{wget}, \texttt{ssh}, \texttt{nginx}, \texttt{apache}, \texttt{iptables}, \texttt{firewall} \\
    File and text processing & \texttt{file-io}, \texttt{tar}, \texttt{gzip}, \texttt{zip}, \texttt{json}, \texttt{csv}, \texttt{regex}, \texttt{xml}, \texttt{string}, \texttt{sed}, \texttt{awk}, \texttt{grep}, \texttt{yaml}, \texttt{parquet} \\
    Python ecosystem & \texttt{python}, \texttt{pip}, \texttt{virtualenv}, \texttt{conda}, \texttt{pandas}, \texttt{numpy}, \texttt{matplotlib}, \texttt{jupyter-notebook}, \texttt{scipy}, \texttt{sympy} \\
    Machine learning and data & \texttt{machine-learning}, \texttt{deep-learning}, \texttt{pytorch}, \texttt{tensorflow}, \texttt{keras}, \texttt{scikit-learn}, \texttt{nlp}, \texttt{huggingface-transformers} \\
    Databases and storage & \texttt{sqlite}, \texttt{mysql}, \texttt{postgresql}, \texttt{redis}, \texttt{mongodb} \\
    Web automation and APIs & \texttt{web-scraping}, \texttt{requests}, \texttt{beautifulsoup}, \texttt{selenium}, \texttt{api} \\
    Security and cryptography & \texttt{security}, \texttt{encryption}, \texttt{cryptography}, \texttt{hash} \\
    Debugging and reliability & \texttt{debugging}, \texttt{logging}, \texttt{exception} \\
    Algorithms and concurrency & \texttt{algorithm}, \texttt{sorting}, \texttt{recursion}, \texttt{data-structures}, \texttt{multiprocessing}, \texttt{multithreading}, \texttt{asyncio} \\
    Media and scientific tools & \texttt{ffmpeg}, \texttt{opencv}, \texttt{image-processing}, \texttt{r} \\
    Build and editor tooling & \texttt{makefile}, \texttt{cmake}, \texttt{gcc}, \texttt{vim}, \texttt{tmux} \\
    \bottomrule
  \end{tabularx}
\end{table}

\subsection{Phase 2: Cascaded LLM Task Generation}

\textbf{Model:} \texttt{claude-opus-4-6}.

\textbf{Generation order:} Strictly cascaded. Each stage receives all prior outputs as context, reducing cross-component inconsistency. The seven stages are: (1) Instruction (temp=0.7), (2) Environment (temp=0.3), (3) Solution (temp=0.3), (4) Difficulty (temp=0.1), (5) Tests (temp=0.2), (6) Test Review (temp=0.1, up to 3 rounds), (7) Dockerfile (temp=0.3).

\textbf{System prompt (shared by all stages):}
\begin{small}
\begin{promptbox}
You are a professional technical task designer responsible for converting
StackOverflow questions into Terminal Bench format programming tasks.
You need to generate clear, executable, and testable tasks. All outputs
should be actual runnable code and scripts.
Please ensure the generated content is in correct format and can be
directly saved as files.
\end{promptbox}
\end{small}

\subsubsection{Stage 1: Instruction Generation (temperature=0.7, max\_tokens=16384)}

\begin{small}
\begin{promptbox}
Please convert the following StackOverflow question into a Terminal Bench
task instruction.md file.

**Original Question:**
Title: {title}
Tags: {tags}
Content:
{body}

**Requirements:**
1. Write a clear task description in Markdown format
2. The task should be completed in a Linux terminal environment
3. Specify clear working directory paths (use /app/task_file/ as root directory)
4. If input files are needed, specify file location (e.g., /app/task_file/input/)
5. If output files are needed, specify output location (e.g., /app/task_file/output/)
6. Provide specific success criteria

Please output the instruction.md content directly in markdown format.
Do NOT wrap in code blocks.
\end{promptbox}
\end{small}

\subsubsection{Stage 2: Environment Generation (temperature=0.3)}

\begin{small}
\begin{promptbox}
Based on the following Terminal Bench task, analyze and generate the
required environment files.

**Task instruction:**
{instruction}

**Original question info:**
Title: {title}
Tags: {tags}

**Requirements:**
1. Analyze what preset files the task needs (e.g., input data, config files)
2. Generate reasonable test data
3. File paths relative to environment/ directory
4. If needed, create subdirectory structure like task_file/input/

**Output format (JSON):**
```json
{
    "files": {
        "relative/path/filename": "file content",
        "task_file/input/example.txt": "example content..."
    },
    "directories": ["task_file", "task_file/input", "task_file/output"]
}
```

Please output only JSON, wrapped with ```json```.
If no files are needed, return empty files and directories.
\end{promptbox}
\end{small}

\subsubsection{Stage 3: Solution Generation (temperature=0.3)}

\begin{small}
\begin{promptbox}
Based on the following Terminal Bench task and StackOverflow answer,
generate solution/solve.sh file.

**Task instruction:**
{instruction}

**StackOverflow answer:**
{answer}

**Tags:** {tags}

**Environment files available in the container:**
{env_file_list}

**Requirements:**
1. Generate an executable bash script
2. The script should complete the task requirements
3. Include necessary comments
4. Handle possible error cases
5. Ensure output meets task requirements
6. The script runs inside the container at WORKDIR /app

**Output format:**
```bash
#!/bin/bash
# Terminal-Bench Canary String, DO NOT REMOVE:
# terminal-bench-canary GUID {guid}

# Your solution code...
```

Please output only bash script content, wrapped with ```bash```.
\end{promptbox}
\end{small}

\subsubsection{Stage 4: Difficulty Classification (temperature=0.1)}

\begin{small}
\begin{promptbox}
Assess the difficulty of the following terminal/programming task for an
experienced software engineer working in a Linux terminal.

**Task instruction:**
{instruction}

**Tags:** {tags}

**Difficulty levels:**
- **easy**: Straightforward single-command or simple script task.
  Most engineers solve it in < 5 minutes.
  Examples: basic file operations, simple grep/find, common git commands.
- **medium**: Requires combining multiple tools or writing a multi-step script.
  Takes 5-15 minutes.
  Examples: parsing structured data, configuring services, writing pipelines.
- **hard**: Requires deeper domain knowledge, complex debugging, or
  multi-component system understanding. Takes > 15 minutes.
  Examples: network troubleshooting, compiler errors, distributed systems.

Output only one word: easy, medium, or hard.
\end{promptbox}
\end{small}

\subsubsection{Stage 5: Test Generation (temperature=0.2)}

\textbf{Key design:} Tests receive the full context (instruction + environment + solution) to ensure consistency. The prompt explicitly forbids re-running \texttt{solve.sh} and requires tests to verify post-execution state only.

\begin{small}
\begin{promptbox}
Based on the following Terminal Bench task, its environment and its
reference solution, generate the test code.

**Task instruction:**
{instruction}

**Environment files in the container:**
{env_file_list}

**Reference solution (solve.sh) that will be executed:**
```bash
{solution}
```

**Tags:** {tags}

**Generate two files:**

1. **test.sh** - Test runner script, format:
```bash
# Terminal-Bench Canary String, DO NOT REMOVE:
# terminal-bench-canary GUID {guid}
#!/bin/bash

apt-get update
apt-get install -y curl

# Install uv
curl -LsSf https://astral.sh/uv/0.9.5/install.sh | sh

source \$HOME/.local/bin/env

# Check if we're in a valid working directory
if [ "\$PWD" = "/" ]; then
    echo "Error: No working directory set."
    exit 1
fi

# Run pytest tests
uvx \
  -p 3.13 \
  -w pytest==8.4.1 \
  -w pytest-json-ctrf==0.3.5 \
  pytest --ctrf /logs/verifier/ctrf.json /tests/test_outputs.py -rA

if [ \$? -eq 0 ]; then
  echo 1 > /logs/verifier/reward.txt
else
  echo 0 > /logs/verifier/reward.txt
fi
```

2. **test_outputs.py** - pytest test file that verifies the state AFTER
solve.sh has already been executed:

**CRITICAL RULES for test_outputs.py:**
- solve.sh has ALREADY been executed before test_outputs.py runs.
  Do NOT call solve.sh again via subprocess.
- Do NOT reference or invoke solve.sh in any test. It does not exist at test time.
- Tests should ONLY check the resulting state: output files, file contents,
  directory structures, installed packages, etc.
- Use os.path.exists(), open().read(), subprocess.run() for verification
  commands (like `grep`, `wc`, `cat`, `ls`), but NEVER run solve.sh.
- Example: if solve.sh creates /app/task_file/output/results.txt, test should
  check that file exists and contains expected content.

**ROBUSTNESS RULES (very important):**
- When checking output files that list paths, NEVER compare exact path strings.
  Instead use `line.endswith("filename")` or `os.path.basename(line)` to handle
  both relative and absolute paths.
- When checking file contents, use `in` operator or regex, NOT exact equality
  (e.g., `assert "expected_text" in content`).
- When counting lines, allow for trailing newlines:
  `len([l for l in content.strip().splitlines() if l.strip()])`.
- When checking numeric results, allow small floating-point tolerance if applicable.
- Prefer checking that files/directories EXIST on disk (os.path.exists,
  os.path.isfile) over parsing output text files.
- If the task creates output files, read them directly with open(); if the task
  modifies existing files, check the modified content.

**Output format (JSON):**
```json
{
    "test_sh": "#!/bin/bash\n...",
    "test_outputs_py": "# Terminal-Bench Canary String...\nimport pytest\n..."
}
```

Please output only JSON, wrapped with ```json```.
Ensure Python code syntax is correct and all tests pass after solve.sh runs.
\end{promptbox}
\end{small}

\subsubsection{Stage 6: Test Review Loop (temperature=0.1)}

After generating tests, the pipeline applies \texttt{ast.parse()} to catch syntax errors in \texttt{test\_outputs.py}, then asks an independent LLM reviewer to inspect the test against five common failure modes. If the review fails, the reported issues are fed back into the next generation round. The max rounds is up to 3.

\begin{small}
\begin{promptbox}
You are reviewing a pytest test file for a Terminal Bench task.
The test runs AFTER solve.sh has already been executed.

**Task instruction:**
{instruction}

**Environment files:**
{env_file_list}

**Reference solution (solve.sh):**
```bash
{solution}
```

**Generated test code (test_outputs.py):**
```python
{test_code}
```

**Review checklist:**
1. Does the test try to run solve.sh via subprocess? (FAIL if yes)
2. Does the test use brittle exact-path comparisons instead of endswith/basename? (WARN)
3. Does the test hardcode values that don't match the solution's actual output? (FAIL)
4. Does the test have missing imports or syntax errors? (FAIL)
5. Do the assertions actually check the expected post-solve.sh state? (FAIL if not)

**Output format (JSON):**
```json
{
    "pass": true,
    "issues": []
}
```
or
```json
{
    "pass": false,
    "issues": ["Issue 1: ...", "Issue 2: ..."]
}
```

Please output only JSON, wrapped with ```json```.
\end{promptbox}
\end{small}

\subsubsection{Stage 7: Dockerfile Generation (temperature=0.3)}

\begin{small}
\begin{promptbox}
Based on the following Terminal Bench task, generate a Dockerfile for the environment.

**Task instruction:**
{instruction}

**Tags:** {tags}

**Requirements:**
1. Choose an appropriate base image based on the task requirements:
   - For Python tasks: use `python:3.13-slim-bookworm`
   - For Node.js tasks: use `node:20-slim`
   - For Java tasks: use `openjdk:17-slim`
   - For Go tasks: use `golang:1.21-bookworm`
   - For general Linux/shell tasks: use `ubuntu:22.04`
2. Install necessary packages for the task
3. Set WORKDIR to /app
4. Include the COPY command to copy task_file into the container
5. The Dockerfile must start with the canary string comment

**Output format:**
```dockerfile
# Terminal-Bench Canary String, DO NOT REMOVE:
# terminal-bench-canary GUID {guid}
FROM <base_image>
WORKDIR /app

RUN apt-get update && apt-get install -y <packages> && rm -rf /var/lib/apt/lists/*

COPY ./task_file /app/task_file
```

Please output only Dockerfile content, wrapped with ```dockerfile```.
\end{promptbox}
\end{small}

\subsection{Phase 3: Docker Round-Trip Validation}

For each candidate task, the validator performs a full Docker lifecycle test:

\begin{enumerate}
    \item \textbf{Build:} \texttt{docker build -t tb2-validate-<task\_name> <task\_dir>/environment}
    \item \textbf{Start container:} \texttt{docker run --name <container\_name> -d <image\_tag> sleep infinity}
    \item \textbf{Copy solution:} \texttt{docker cp <task\_dir>/solution/solve.sh <container>:/app/}
    \item \textbf{Execute solution:} \texttt{docker exec <container> bash /app/solve.sh}
    \item \textbf{Copy tests:} \texttt{docker cp <task\_dir>/tests <container>:/tests}
    \item \textbf{Run tests:} \texttt{docker exec <container> bash /tests/test.sh}
    \item \textbf{Read reward:} \texttt{docker cp <container>:/logs/verifier/reward.txt <local>}
    \item \textbf{Check:} If \texttt{reward.txt} contains ``1'', copy task to output directory; otherwise skip.
    \item \textbf{Cleanup:} \texttt{docker stop <container> \&\& docker rm <container> \&\& docker rmi <image\_tag>}
\end{enumerate}

\textbf{Parallelization:} 8 workers by default. Each worker processes one task at a time to avoid Docker resource contention.

\textbf{Timeout:} 300 seconds per task (configurable via \texttt{--timeout}). Tasks that exceed the timeout are marked as ``timeout'' and skipped.

\textbf{Error handling:} Build failures, runtime errors, missing solution files, and test failures are logged separately. Only tasks with \texttt{reward=1} are retained.

\subsection{Phase 4: Dataset Packaging}

Renumbers validated tasks as \texttt{task\_00000}, \texttt{task\_00001}, \ldots and copies them to the final output directory. Generates \texttt{generation\_summary.json} with difficulty distribution, category distribution, and task mapping.

\textbf{Docker validation pass rate:} 15,389 / 36,846 = 41.8\%. The main failure modes are test/solution mismatch (24.0\%), build failures (11.5\%), and timeouts (22.6\%).

\begin{table}[t]
  \caption{Validation funnel for Terminal-Lego task construction. The final passed set contains Docker round-trip verified tasks used for trajectory collection.}
  \label{tab:dataset-validation-funnel}
  \centering
  \small
  \begin{tabular}{lr}
    \toprule
    \textbf{Outcome} & \textbf{Estimated Count} \\
    \midrule
    Test/solution failed & 8,859 \\
    Build failed & 4,254 \\
    Timeout & 8,317 \\
    Runtime error & 18 \\
    Missing solution & 9 \\
    Passed & 15,389 \\
    \midrule
    Total candidates validated & 36,846 \\
    \bottomrule
  \end{tabular}
\end{table}

\subsubsection{Test Pass Rate}

After constructing the test file, we run the oracle solution and test the \texttt{FAIL\_TO\_PASS} and \texttt{PASS\_TO\_PASS}. The results are shown in Table.~\ref{tab:f2p-p2p-tests}. The datasets contains substantially more fail-to-pass tests than pass-to-pass tests, indicating that the evaluation is primarily focused on verifying whether solutions resolve the target issue.

\begin{table}[h]
\centering
\caption{
  Statistics of fail-to-pass and pass-to-pass tests.
}
\label{tab:f2p-p2p-tests}
\small
\begin{tabular}{lc}
\toprule
\textbf{Test Type} & \textbf{Number of Tests}\\
\midrule
Fail-to-Pass & 8.81  \\
Pass-to-Pass & 1.48 \\
\bottomrule
\end{tabular}
\end{table}

\section{Terminus-2 Agent Scaffold and Interaction Protocol}
\label{sec:terminus2}

Terminus-2 is a minimal, unopinionated agent scaffold developed by the Terminal-Bench team as a neutral testbed for comparing language model performance~\cite{merrill2026terminal}. Because Terminal-Bench is an interactive framework, agent and model performance are hard to decouple---many agent scaffolds have been engineered to accommodate the tendencies of certain models, especially when the model and agent are developed by the same organization. Terminus-2 is designed to eliminate this confound.

\paragraph{Single-tool design.}
Whereas other agents (e.g., Claude Code, OpenHands, Gemini CLI) provide dedicated tools for editing files, executing Bash commands, or downloading files, Terminus-2 operates solely by issuing Bash commands through a headless terminal. The agent runs in a simple loop: a language model produces keystrokes, which are executed in a \texttt{tmux} shell inside a Docker container; after execution, the captured terminal output is returned to the model as the next observation. This single-tool constraint gives the model full flexibility to decide \emph{how} to accomplish sub-tasks---it may echo contents directly into a file, operate an interactive text editor like \texttt{vim}, scroll through output, use arrow keys to navigate a menu, or launch additional shells.

\paragraph{Context summarization.}
Terminus-2 incorporates a context summarization module that condenses the agent's conversation history when it approaches the model's context limit. The module invokes the language model itself to produce a structured handoff summary, enabling the agent to operate on extended tasks that may require millions of tokens without losing critical information from earlier episodes.

\paragraph{Interaction protocol.}
Each turn, the model produces a structured JSON response containing three required fields (\texttt{analysis}, \texttt{plan}, \texttt{commands}) and one optional field (\texttt{task\_complete}). Each command object specifies verbatim \texttt{keystrokes} to send to the terminal and a \texttt{duration} (seconds to wait before capturing output). Task completion requires double confirmation---the agent must set \texttt{task\_complete: true} in two consecutive turns before the trial is finalized, reducing premature termination.

\paragraph{Architecture overview.}
Table~\ref{tab:terminus2} summarizes the key components of Terminus-2.

\begin{table}[h]
\centering
\caption{Terminus-2 agent architecture summary.}
\label{tab:terminus2}
\begin{tabular}{ll}
\toprule
\textbf{Component} & \textbf{Description} \\
\midrule
Tool & Single headless terminal (\texttt{tmux} session in Docker) \\
Execution & Bash commands only; keystrokes sent verbatim \\
LLM Backend & LiteLLM (model-agnostic; supports vLLM, OpenAI-compatible APIs) \\
Output Format & Structured JSON (\texttt{analysis}, \texttt{plan}, \texttt{commands}, \texttt{task\_complete}) \\
Output Truncation & 10\,000 bytes; head/tail split preserving boundary information \\
Context Management & Proactive summarization via Q\&A-based handoff protocol \\
Completion & Double confirmation (two consecutive \texttt{task\_complete: true}) \\
Error Recovery & Retry up to 3$\times$; context unwind + re-summarize on overflow \\
\bottomrule
\end{tabular}
\end{table}

\paragraph{Role in Terminal-Bench evaluation.}
Terminus-2 serves as the standardized baseline throughout the Terminal-Bench 2.0 evaluation:
\begin{itemize}
  \item \textbf{Empirical difficulty} of each task is defined by Terminus-2's average pass rate across frontier models (Easy $\geq$66.7\%, Medium 33.3--66.7\%, Hard $<$33.3\%).
  \item \textbf{Trajectory-level error analysis} uses Terminus-2 as the fixed scaffold to isolate model-level differences.
  \item \textbf{Command-level error analysis} reviews individual command input-output pairs from recorded Terminus-2 trajectories.
\end{itemize}


\section{Training Dynamics Across Teachers} \label{sec:appendix_training_dynamics}

This section compares the SFT training dynamics when fine-tuning Qwen3-32B on trajectories from four different teachers. All experiments use identical hyperparameters (learning rate, batch size, warmup, scheduler) and train for the same number of effective steps, differing only in the source of training trajectories.

\subsection{Training Loss  Across Teacher Trajectories}

\begin{figure}[h] 
\centering
\includegraphics[width=0.85\textwidth]{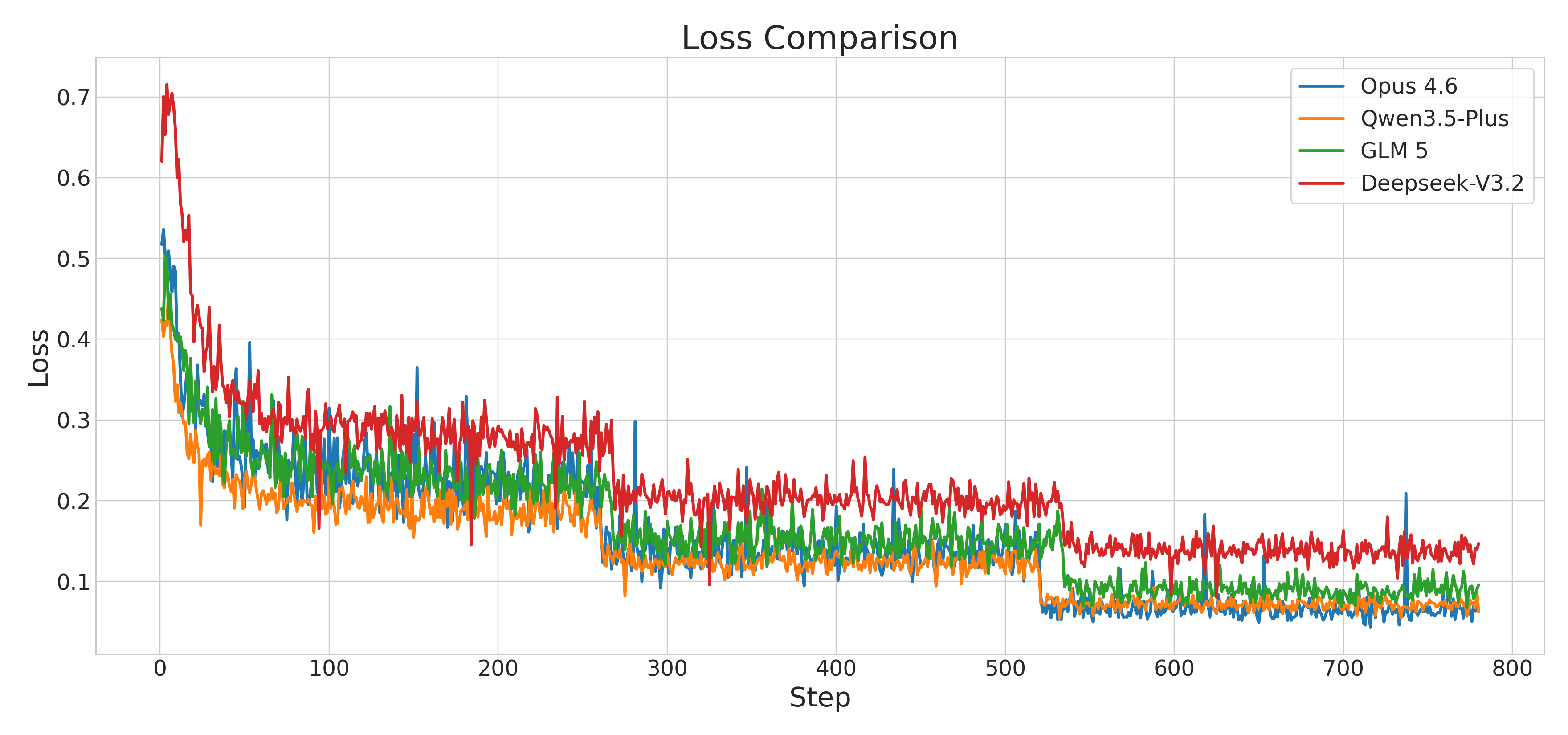} 
\caption{Training loss curves when fine-tuning Qwen3-32B on trajectories from different teachers. DeepSeek-V3.2 trajectories yield significantly higher loss than other teachers throughout training.} \label{fig:training_loss}
\end{figure}

Figure~\ref{fig:training_loss} shows the training loss curves for each teacher's trajectories.

\begin{table}[h] 
\centering 
\caption{Training loss statistics across teacher trajectories.} 
\label{tab:training_loss} 
\begin{tabular}{lrrr} 
\toprule 
\textbf{Teacher} & \textbf{Final Loss} & \textbf{Avg Last 10\%} & \textbf{Min Loss} \\ 
\midrule DeepSeek-V3.2 & 0.1831 & 0.1773 & 0.1264 \\
GLM-5 & 0.0703 & 0.0726 & 0.0552 \\
Qwen3.5-Plus & 0.0627 & 0.0713 & 0.0544 \\
Claude Opus 4.6 & 0.0631 & 0.0661 & 0.0435 \\
\bottomrule 
\end{tabular} 
\end{table}

DeepSeek-V3.2 trajectories produce a final loss of 0.183---approximately 2.6$\times$ higher than the other three teachers (0.063--0.070). This gap persists throughout training and does not close with additional steps, indicating a fundamental distributional mismatch between DeepSeek-V3.2's behavioral patterns and Qwen3-32B's prior. The remaining three teachers converge to similar loss levels, with Qwen3.5-Plus trajectories (self-overlap) achieving the lowest loss as expected.

\subsection{Gradient Norm Across Teacher Trajectories}

\begin{figure}[h] 
\centering 
\includegraphics[width=0.85\textwidth]{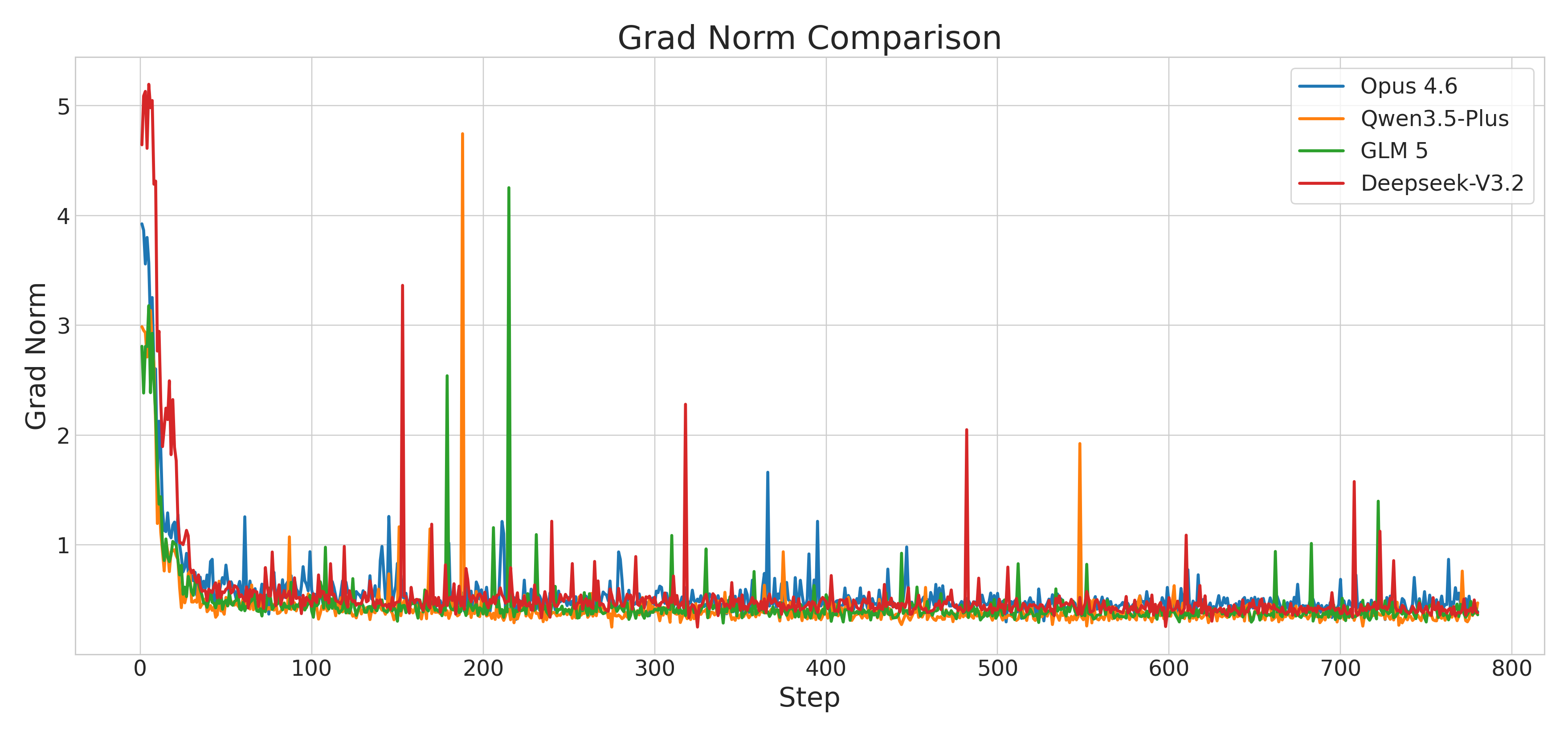} 
\caption{Gradient norm curves during fine-tuning. Claude Opus 4.6 exhibits the highest gradient variance despite low loss, while DeepSeek-V3.2 shows the most stable gradients.}
\label{fig:training_gradnorm}
\end{figure}

Figure~\ref{fig:training_gradnorm} shows the gradient norm curves during training.

\begin{table}[h] 
\centering 
\caption{Gradient norm statistics across teacher trajectories. We report the average gradient norm over the last 10\% of training steps.} 
\label{tab:training_gradnorm}
\begin{tabular}{lr} 
\toprule 
\textbf{Teacher} & \textbf{Avg Last 10\% Grad Norm} \\ 
\midrule 
DeepSeek-V3.2 & 0.2965 \\
GLM-5 & 0.3724 \\ 
Qwen3.5-Plus & 0.3603 \\
Claude Opus 4.6 & 0.4598 \\
\bottomrule 
\end{tabular} 
\end{table}

An inverse relationship emerges between loss and gradient stability: DeepSeek-V3.2, despite having the highest loss, exhibits the lowest gradient norm (0.30), indicating that the model has settled into a stable plateau---it cannot further reduce the loss but is not oscillating. Conversely, Claude Opus 4.6 achieves low loss but with the highest gradient norm (0.46), suggesting that the optimization landscape for Opus trajectories contains sharper local structure.

\subsection{Imitation Difficulty and Downstream Performance}

Table~\ref{tab:imitation_difficulty} summarizes the relationship between imitation difficulty (measured by training loss) and downstream task performance.

\begin{table}[h] 
\centering 
\caption{Imitation difficulty vs. downstream performance. Teachers are ordered by training loss (imitation difficulty).} 
\label{tab:imitation_difficulty} 
\begin{tabular}{llll} 
\toprule 
\textbf{Teacher} & \textbf{Imitation Difficulty} & \textbf{Training Loss} & \textbf{Gradient Stability}\\
\midrule 
Qwen3.5-Plus & Easiest & 0.063 & Stable (0.36)\\ 
Claude Opus 4.6 & Easy & 0.063 & Unstable (0.46) \\ 
GLM-5 & Moderate & 0.070 & Stable (0.37) \\
DeepSeek-V3.2 & Hardest & 0.183 & Most stable (0.30) \\
\bottomrule 
\end{tabular} 
\end{table}

\paragraph{The imitation--performance paradox.} 
The teacher whose trajectories are hardest to imitate (DeepSeek-V3.2, loss 0.183) produces the best downstream task performance, while the easiest-to-imitate teachers (Qwen3.5-Plus, Claude Opus 4.6) yield weaker results. This paradox resolves when we consider \emph{what} makes DeepSeek-V3.2 hard to imitate: its distinctive observation-heavy behavioral paradigm---characterized by high targeted observation rates and systematic pre-action reconnaissance---diverges substantially from Qwen3-32B's default behavior. The high loss reflects not training failure, but the magnitude of behavioral shift required.

\paragraph{Gradient stability as a convergence signal.} 
DeepSeek-V3.2's low gradient norm (0.30) combined with high loss indicates that the model has converged to a stable representation of the teacher's behavior, even though it cannot perfectly reproduce it. The residual loss captures the irreducible gap between the student's capacity and the teacher's behavioral complexity. In contrast, Claude Opus 4.6's high gradient norm (0.46) despite low loss suggests that the model oscillates between multiple low-loss solutions without settling on a consistent behavioral pattern.

\paragraph{Implications for teacher selection.} 
These results suggest that training loss alone is a poor predictor of downstream performance. A teacher whose behavioral paradigm is \\emph{complementary} to the student's prior---introducing novel patterns such as systematic observation---provides greater learning signal than a teacher whose behavior the student can already approximate. The difficulty of imitation is itself evidence that the teacher offers something the student lacks.

\section{Observation-Action Patterns}
\label{sec:appendix_oa_patterns}

This section provides a detailed analysis of how observation commands influence subsequent actions in teacher trajectories. All statistics are computed on the DeepSeek-V3.2 trajectory corpus (15{,}389 samples, 120{,}919 assistant turns, 93{,}287 turns containing at least one parsed command). We distinguish between \texttt{cat} used for reading (\texttt{cat file}, classified as observation) and \texttt{cat} used for writing (\texttt{cat > file}, \texttt{cat << EOF}, classified as action), as these represent fundamentally different behaviors despite sharing a command name.

\subsection{Turn-Level Classification}

We classify each assistant turn by the commands it contains:

\begin{table}[h]
\centering
\caption{Turn-level classification of assistant turns containing commands.}
\label{tab:turn_classification_oa}
\begin{tabular}{lrr}
\toprule
\textbf{Turn Type} & \textbf{Count} & \textbf{\%} \\
\midrule
Mixed (obs + action) & 44{,}768 & 48.0\% \\
Observation-only & 22{,}245 & 23.8\% \\
Action-only & 21{,}809 & 23.4\% \\
Other & 4{,}465 & 4.8\% \\
\midrule
\textbf{Total} & \textbf{93{,}287} & \textbf{100\%} \\
\bottomrule
\end{tabular}
\end{table}

Among turns that contain at least one parsed command, 71.8\% (67{,}013 / 93{,}287) include an observation command. Mixed turns---where observation and action co-occur---account for 48.0\% of all command-bearing turns. Notably, once \texttt{cat}-write is properly classified as action, observation-only and action-only turns appear in near-equal proportion (23.8\% vs.\ 23.4\%), revealing a balanced interleaving of information gathering and environment modification.

\subsection{Observation Command Taxonomy}

We classify \texttt{cat file} (reading) as observation and \texttt{cat > file} / \texttt{cat << EOF} (writing) as action. Of 88{,}208 total \texttt{cat} occurrences in the corpus, 43{,}330 (49.1\%) are reads and 44{,}878 (50.9\%) are writes.

Table~\ref{tab:obs_cmd_freq} shows the frequency of observation commands.

\begin{table}[h]
\centering
\caption{Observation command frequency across 138{,}488 total observation command occurrences.}
\label{tab:obs_cmd_freq}
\begin{tabular}{lrrr}
\toprule
\textbf{Command} & \textbf{Count} & \textbf{\% of Total} & \textbf{Purpose} \\
\midrule
\texttt{ls} & 47{,}236 & 34.1\% & List directory structure \\
\texttt{cat} (read) & 43{,}330 & 31.3\% & Read file contents \\
\texttt{echo} & 14{,}790 & 10.7\% & Display text / test output \\
\texttt{grep} & 8{,}720 & 6.3\% & Search file contents \\
\texttt{head} & 6{,}639 & 4.8\% & Preview file start \\
\texttt{wc} & 4{,}650 & 3.4\% & Count lines/words/chars \\
\texttt{which} & 4{,}258 & 3.1\% & Locate executables \\
\texttt{find} & 3{,}258 & 2.4\% & Search file system \\
\texttt{diff} & 1{,}682 & 1.2\% & Compare files \\
\texttt{tail} & 1{,}275 & 0.9\% & Preview file end \\
\bottomrule
\end{tabular}
\end{table}

Two commands dominate: \texttt{ls} (34.1\%) and \texttt{cat}-read (31.3\%), together accounting for 65.4\% of all observation activity. Their near-parity reflects the two fundamental questions an agent asks: ``what files exist?'' (structure inspection) and ``what does the file contain?'' (content inspection). Agents balance structural navigation with content-level understanding, rather than strongly favoring one over the other.

\subsection{Action Command Distribution}

Table~\ref{tab:act_cmd_freq} shows the frequency of action commands, with \texttt{cat}-write (heredoc file creation) classified as action.

\begin{table}[h]
\centering
\caption{Action command frequency across 153{,}496 total action command occurrences.}
\label{tab:act_cmd_freq}
\begin{tabular}{lrrr}
\toprule
\textbf{Command} & \textbf{Count} & \textbf{\% of Total} & \textbf{Purpose} \\
\midrule
\texttt{cat} (write) & 44{,}878 & 29.2\% & Write/create file contents \\
\texttt{python3} & 28{,}079 & 18.3\% & Execute Python scripts \\
\texttt{git} & 17{,}796 & 11.6\% & Version control operations \\
\texttt{mkdir} & 11{,}608 & 7.6\% & Create directories \\
\texttt{chmod} & 10{,}784 & 7.0\% & Modify file permissions \\
\texttt{echo} (write) & 6{,}942 & 4.5\% & Write text to file \\
\texttt{rm} & 4{,}550 & 3.0\% & Remove files \\
\texttt{python} & 4{,}444 & 2.9\% & Execute Python scripts \\
\texttt{sed} & 3{,}387 & 2.2\% & Stream editing \\
\texttt{node} & 3{,}332 & 2.2\% & Execute JavaScript \\
\texttt{cp} & 2{,}668 & 1.7\% & Copy files \\
\texttt{bash} & 2{,}547 & 1.7\% & Execute shell scripts \\
\bottomrule
\end{tabular}
\end{table}

File creation via \texttt{cat}-write (29.2\%) is the single most frequent action command, reflecting the dominant pattern of using heredocs (\texttt{cat << 'EOF' > file}) to create or overwrite files. Together with \texttt{echo}-write (4.5\%), file-writing operations account for 33.7\% of all actions. Script execution (\texttt{python3} + \texttt{python}, 21.2\%) is the second major category, followed by file system operations (\texttt{mkdir}, \texttt{chmod}, \texttt{rm}, \texttt{cp}: 19.3\%). The dominance of \texttt{cat}-write over \texttt{sed} (2.2\%) indicates that agents overwhelmingly prefer creating complete files over in-place editing.

\subsection{Observation$\to$Action Pairing Patterns}

We analyze how observation commands relate to subsequent actions by tracking command pairs within a window of 3 assistant turns. For each observation command in a turn, we identify the action commands that follow within the window.

\paragraph{Overall pairing statistics.}
Across 15{,}389 trajectories, we identified 490{,}299 observation$\to$action pairs within a 3-turn window. The high volume reflects the dense interleaving of observation and action throughout agent trajectories.

\paragraph{Most frequent observation$\to$action pairs.}

Table~\ref{tab:oa_pairs} shows the most common observation$\to$action command pairs.

\begin{table}[h]
\centering
\caption{Top observation$\to$action pairs within a 3-turn window (490{,}299 total pairs).}
\label{tab:oa_pairs}
\begin{tabular}{llrr}
\toprule
\textbf{Purpose} & \textbf{Pair} & \textbf{Count} & \textbf{\%} \\
\midrule
Structure $\to$ create file & \texttt{ls}$\to$\texttt{cat}-write & 53{,}582 & 10.9\% \\
Inspect $\to$ create file & \texttt{cat}$\to$\texttt{cat}-write & 48{,}912 & 10.0\% \\
Structure $\to$ execute & \texttt{ls}$\to$\texttt{python3} & 31{,}728 & 6.5\% \\
Inspect $\to$ execute & \texttt{cat}$\to$\texttt{python3} & 28{,}266 & 5.8\% \\
Inspect $\to$ version control & \texttt{cat}$\to$\texttt{git} & 19{,}954 & 4.1\% \\
Structure $\to$ version control & \texttt{ls}$\to$\texttt{git} & 19{,}416 & 4.0\% \\
Structure $\to$ create dir & \texttt{ls}$\to$\texttt{mkdir} & 16{,}108 & 3.3\% \\
Structure $\to$ set perms & \texttt{ls}$\to$\texttt{chmod} & 15{,}383 & 3.1\% \\
Inspect $\to$ set perms & \texttt{cat}$\to$\texttt{chmod} & 12{,}548 & 2.6\% \\
Inspect $\to$ write text & \texttt{cat}$\to$\texttt{echo}-write & 11{,}613 & 2.4\% \\
\bottomrule
\end{tabular}
\end{table}

The two dominant patterns are \textbf{structure inspection$\to$file creation} (\texttt{ls}$\to$\texttt{cat}-write, 10.9\%) and \textbf{content inspection$\to$file creation} (\texttt{cat}$\to$\texttt{cat}-write, 10.0\%). Together, these account for 20.9\% of all pairs, revealing the core workflow: agents inspect the current state (either directory structure or file contents), then create or overwrite files based on what they observe. The \texttt{ls}$\to$\texttt{cat}-write pattern captures the ``check what exists, then create what's missing'' workflow, while \texttt{cat}$\to$\texttt{cat}-write captures ``read the current version, then write an updated version.''

\subsection{Turn Transition Patterns}

Table~\ref{tab:turn_transitions_oa} shows the bigram distribution of consecutive assistant turn types, revealing how observation and action behaviors chain across turns.

\begin{table}[h]
\centering
\caption{Consecutive assistant turn type transitions (top patterns).}
\label{tab:turn_transitions_oa}
\begin{tabular}{llrr}
\toprule
\textbf{Turn $t$} & \textbf{Turn $t{+}1$} & \textbf{Count} & \textbf{\%} \\
\midrule
Mixed & Mixed & 21{,}089 & 26.9\% \\
Action-only & Action-only & 9{,}147 & 11.7\% \\
Mixed & Action-only & 7{,}985 & 10.2\% \\
Mixed & Obs-only & 7{,}280 & 9.3\% \\
Action-only & Mixed & 7{,}267 & 9.3\% \\
Obs-only & Mixed & 7{,}137 & 9.1\% \\
Obs-only & Obs-only & 4{,}563 & 5.8\% \\
Action-only & Obs-only & 3{,}259 & 4.2\% \\
\bottomrule
\end{tabular}
\end{table}

The most common transition is Mixed$\to$Mixed (26.9\%), indicating sustained interleaving of observation and action. The Action-only$\to$Action-only transition (11.7\%) captures consecutive file-creation turns where agents write multiple files in sequence. The Mixed$\to$Action-only transition (10.2\%) shows a common pattern where initial inspection (mixed turn) is followed by confident execution (action-only turn). The Action-only$\to$Obs-only transition (4.2\%) captures post-action verification: after modifying the environment, agents dedicate a turn to inspecting the results.

\subsection{Observation Positioning Within Mixed Turns}

In mixed turns (where both observation and action commands appear), we analyze the \emph{temporal ordering} of observations relative to actions.

\begin{table}[h]
\centering
\caption{Observation positioning within mixed turns (44{,}768 total).}
\label{tab:obs_positioning}
\begin{tabular}{lrr}
\toprule
\textbf{Pattern} & \textbf{Count} & \textbf{\%} \\
\midrule
Observation after action only & 21{,}482 & 48.0\% \\
Observation before action only & 13{,}478 & 30.1\% \\
Observation both before and after & 6{,}706 & 15.0\% \\
Other (interleaved) & 3{,}102 & 6.9\% \\
\bottomrule
\end{tabular}
\end{table}

\paragraph{Post-action verification dominates (48.0\%).}
The most common pattern is observation-after-action, where the agent executes a command and then inspects its output or side effects. This reflects the workflow of writing a file (\texttt{cat}-write) and then verifying its contents (\texttt{cat}-read or \texttt{ls}). With \texttt{cat}-write properly classified as action, this verification-dominant pattern emerges clearly.

\paragraph{Pre-action reconnaissance (30.1\%).}
Observation-before-action captures the information-gathering approach: understand the environment, then act. This is the second most common pattern, indicating that agents frequently inspect the current state before modifying it.

\paragraph{Bracket pattern (15.0\%).}
Observation-both-sides forms a ``sandwich'' pattern: inspect, act, verify. This provides the strongest correctness signal, as the agent can compare pre- and post-action states.

\subsection{First Turn Behavior}

We examine the first assistant turn in each trajectory to understand how agents initialize their problem-solving process.

\begin{table}[h]
\centering
\caption{First turn type distribution across 15{,}389 trajectories.}
\label{tab:first_turn}
\begin{tabular}{lrr}
\toprule
\textbf{Turn Type} & \textbf{Count} & \textbf{\%} \\
\midrule
Mixed (obs + action) & 8{,}270 & 55.8\% \\
Observation-only & 5{,}936 & 40.0\% \\
Action-only & 593 & 4.0\% \\
Other & 32 & 0.2\% \\
\bottomrule
\end{tabular}
\end{table}

95.8\% of trajectories begin with observation (either obs-only or mixed). Only 4.0\% start with action-only turns, indicating that agents almost universally perform reconnaissance before taking their first action. This aligns with the principle of \emph{observe before acting}: successful agents gather information about the task environment before attempting modifications.

\subsection{Key Insights  on Observation-Action Behavior}

\paragraph{Observation is pervasive but not overwhelming.}
71.8\% of command-bearing turns include at least one observation command, and 48.0\% of turns are mixed (obs + action). Once \texttt{cat}-write is properly classified as action, observation-only and action-only turns appear in near-equal proportion (23.8\% vs.\ 23.4\%), revealing a balanced rhythm between information gathering and environment modification.

\paragraph{File creation dominates action behavior.}
\texttt{cat}-write (29.2\%) is the single most frequent action command, indicating that agents primarily modify the environment through complete file creation rather than in-place editing (\texttt{sed}, 2.2\%). This ``write whole files'' strategy is consistent with the heredoc pattern (\texttt{cat << 'EOF' > file}) that allows agents to produce complete, correct files in a single operation.

\paragraph{Post-action verification is the primary observation role in mixed turns.}
Within mixed turns, 48.0\% place observation after action (verification), compared to 30.1\% that place observation before action (reconnaissance). This reversal from naive expectation reflects the dominance of the ``write then verify'' workflow: agents create a file and immediately inspect it to confirm correctness.

\paragraph{Structure and content inspection are balanced.}
\texttt{ls} (34.1\%) and \texttt{cat}-read (31.3\%) contribute nearly equally to observation behavior. Agents balance understanding \emph{what files exist} with understanding \emph{what files contain}, rather than strongly favoring one inspection mode.

\paragraph{The inspect$\to$create workflow dominates cross-turn patterns.}
The two most frequent observation$\to$action pairs are \texttt{ls}$\to$\texttt{cat}-write (10.9\%) and \texttt{cat}$\to$\texttt{cat}-write (10.0\%), together accounting for 20.9\% of all pairs. This reveals the core agent workflow: inspect the current state, then create or overwrite files based on what was observed.

\paragraph{Distinguishing \texttt{cat}-read from \texttt{cat}-write is essential.}
Of 88{,}208 total \texttt{cat} occurrences, 49.1\% are reads (observation) and 50.9\% are writes (action). Treating all \texttt{cat} as observation would inflate observation statistics by 32\% and obscure the true balance between information gathering and environment modification. This distinction is critical for accurate behavioral analysis of coding agents.


\end{document}